\documentclass[10pt,a4paper]{article}
\usepackage[utf8]{inputenc}
\usepackage[english]{babel}
\usepackage{amsmath}
\usepackage{amsfonts}
\usepackage{graphicx}
\usepackage{amssymb}
\usepackage{authblk}
\usepackage{hyperref}
\usepackage{tikz,pgfplots}

\DeclareMathOperator*{\argmax}{arg\,max}
\newcommand{\eg}{\textit{e.g. }}
\newcommand{\ie}{\textit{i.e. }}

\title{Simultaneous Recognition and Pose Estimation of Instruments in Minimally Invasive Surgery}

\author[1]{Thomas Kurmann}
\author[2]{Pablo Marquez Neila}
\author[3]{Xiaofei Du}
\author[2]{Pascal Fua}
\author[3]{Danail Stoyanov}
\author[4]{Sebastian Wolf}
\author[1]{Raphael Sznitman}
\affil[1]{University of Bern, Switzerland}
\affil[2]{École Polytechnique Fédérale de Lausanne, Switzerland}
\affil[3]{University College London, United Kingdom}
\affil[4]{Bern University Hospital, Switzerland}

\date{}

\begin{document}

\maketitle

\begin{abstract}
Detection of surgical instruments plays a key role in ensuring patient safety in minimally invasive surgery. In this paper, we present a novel method for 2D vision-based recognition and pose estimation of surgical instruments that generalizes to different surgical applications. At its core, we propose a novel scene model in order to simultaneously recognize multiple instruments as well as their parts. We use a Convolutional Neural Network architecture to embody our model and show that the cross-entropy loss is well suited to optimize its parameters which can be trained in an end-to-end fashion. An additional advantage of our approach is that instrument detection at test time is achieved while avoiding the need for scale-dependent sliding window evaluation. This allows our approach to be relatively parameter free at test time and shows good performance for both instrument detection and tracking. We show that our approach surpasses state-of-the-art results on {\it in-vivo} retinal microsurgery image data, as well as {\it ex-vivo} laparoscopic sequences.
\end{abstract}

\section{Introduction}\label{sec:intro}
Vision-based detection of surgical instruments in both minimally invasive surgery and microsurgery has gained increasing popularity in the last decade. This is largely due to the potential it holds for more accurate guidance of surgical robots such as the da Vinci\textregistered (Intuitive Surgical, USA) and Preceyes (Netherlands), as well as for directing imaging technology such as endoscopes~\cite{Wolf11} or OCT imaging~\cite{Alsheakhali2016} at manipulated regions of the workspace.

In recent years, a large number of methods have been proposed to either track instruments over time or detect them without any prior temporal information, in both 2D and 3D. In this work, we focus on 2D detection of surgical instruments as it is often required for tracking in both 2D~\cite{Rieke2016} and 3D~\cite{Du2016}. In this context,~\cite{Reiter12,Sznitman2014} proposed to build ensemble-based classifiers using hand-crafted features to detect instruments parts (\eg shaft, tips or center). Similarly, ~\cite{Bouget2015} detected multiple instruments in neurosurgery by repeatedly evaluating a boosted classifier based on semantic segmentation.

Yet for most methods described above two important limitations arise. The first is that instrument detection and pose estimation (\ie instrument position, orientation and location of parts) have been tackled in two phases, leading to complicated pipelines that are sensitive to parameter tuning. The second is that at evaluation time, detection of instruments has been achieved by repeated window sliding at limited scales which is both inefficient and error prone (\eg small or very large instruments are missed). Both points heavily reduce the usability of proposed methods.

In order to overcome these limitations, we propose a novel framework that avoids these and can be applied to a variety of surgical settings. Assuming a known maximum number of instruments and parts that could appear in the field of view, our approach, which relies on recent deep learning strategies~\cite{Ronneberger2015}, avoids the need for window sliding at test time and estimates multiple instruments and their pose simultaneously. This is achieved by designing a novel Convolutional Neural Network (CNN) architecture that explicitly models object parts and the different objects that may be present. We show that when combined with a cross-entropy loss function, our model can be trained in an end-to-end fashion, thus bypassing the need for traditional two-stage detection and pose estimation. We validate our approach on both {\it ex-vivo} laparoscopy images and on {\it in-vivo} retinal microsurgery, where we show improved results over existing detection and tracking methods.


\section{Multi-instrument detector}\label{sec:methods}
\begin{figure}[t!]
\centering
\includegraphics[width=1.0\textwidth]{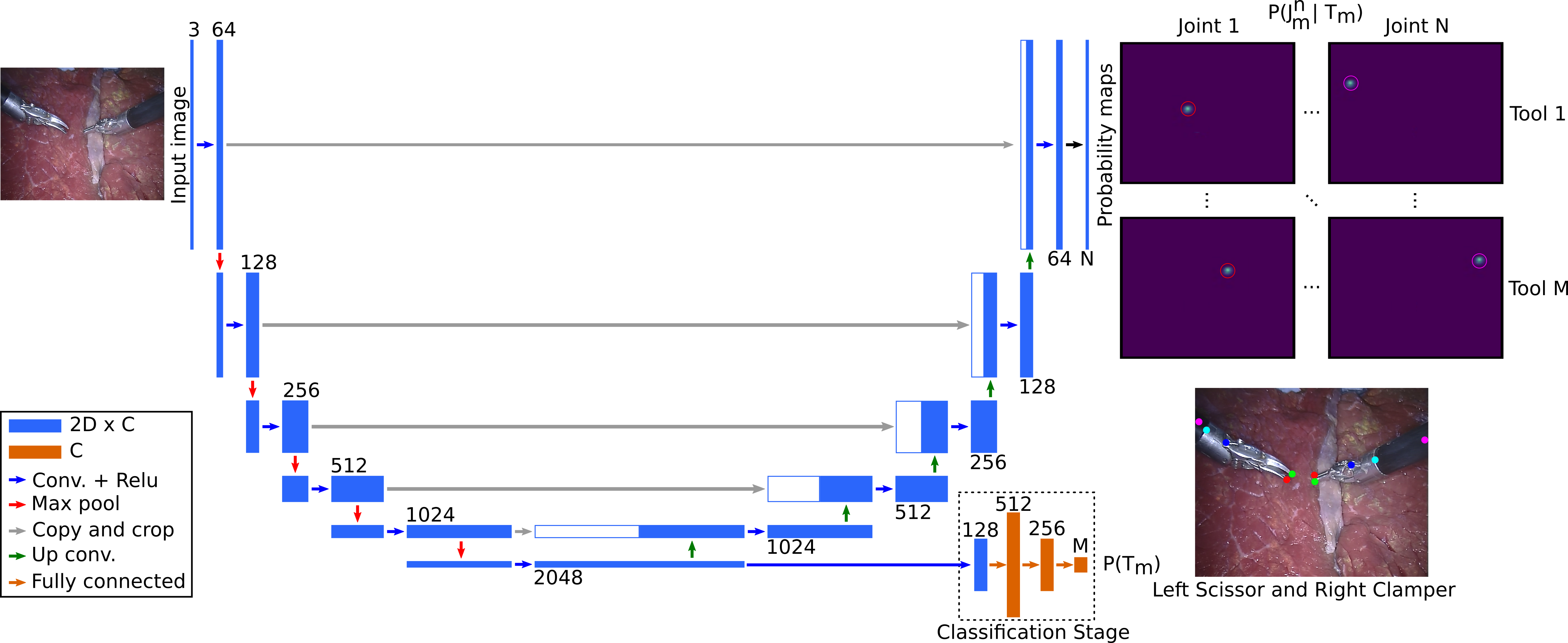}
\caption{Proposed multi-instrument detector network architecture. The network produces probabilistic outputs for both the presence of different instruments and position of their joints. The number of channels C is denoted on top of the box.}
\label{fig:unet}
\end{figure}

In order to detect multiple instruments and their parts in a coherent and simple manner, we propose a scene model which assumes that we know what would be the maximum number of instruments in the field of view. We use a CNN to embody this model and use the cross-entropy to learn effective parameters using a training set. Our CNN architecture takes as input an image and provides binary outputs as to whether or not a given instrument is present as well as 2D location estimates for its parts. A visualization of our proposed detection framework can be seen in Fig.~\ref{fig:unet}. Conveniently then, detecting instruments and estimating the joint positions on a test frame is simply achieved by a feed forward pass of the network. We now describe our scene model and our CNN in more detail.

\subsection{Scene Model}
Let $I \in\mathbb{R}^{w \times h}$ be an image that may contain up to $M$ instruments. In particular, we denote $T=\{T_1,..,T_M\}, T_m \in\{0,1\}$ to be the set of instruments that could appear in the field of view such that $T_{m}=0$ if the tool is not present and $T_{m}=1$ if it is. In addition, each instrument present in the image is defined as a set of $N$ parts, or {\it joints}, $\{J^n_m \in \mathbb{R}^2\}_{n=0}^N$ consisting of 2D image locations. Furthermore, let $GT^{n}_{m} \in \mathbb{R}^2$ be the ground truth 2D position for joint $n$ of instrument $T_m$ and $t_m \in\{0,1\}$ be the ground truth variable indicating if the $mth$ instrument is visible in the image. Assuming that the instrument presence is unknown and is probabilistic in nature, our goal is to train a network to estimate the following scene model
\begin{equation}
P(T_1,\ldots,T_M, J^1_1,\ldots ,J^N_1,\ldots,J^1_M,\ldots,J^N_M)=\prod^{m} P(T_{m}) \prod^{m} \prod^{n} P(J_{m}^{n} | T_{m})
\label{eq:model_1}
\end{equation}
\noindent
where $P(T_{m})$ are Bernoulli random variables and the likelihood models $P(J_{m}^{n} | T_{m})$ are parametric probability distributions. Note that Eq.~\eqref{eq:model_1} assumes independence between the different instruments as well as a conditional independence between the various joints for a given instrument. Even though both assumptions are quite strong, they provide a convenient decomposition and a model simplification of what would otherwise be a complicated distribution.
Letting ${P}$ be the predicted distribution by our CNN and $\hat{P}$ be a probabilistic interpretation of the  ground truth, then the cross-entropy loss function can be defined as
\begin{equation}
 H(\hat{P}, P)=-\sum_{s \in \mathcal{S}} \hat{P}(s)\log P(s)
 \label{eq:loss}
\end{equation}
\noindent
where $\mathcal{S}$ is the probability space over all random variables $(T_1,\ldots,T_M$, $J^1_1$, $\ldots$, $J^N_1,\ldots,J^1_M,\ldots,J^N_M)$. Replacing $P$ and $\hat{P}$ in Eq.~\eqref{eq:loss} with the model Eq.~\eqref{eq:model_1} and simplifying the term gives rise to
\begin{equation}
\begin{split}
H(\hat{P},P) &= \sum_{m} H\left(\hat{P}(T_{m}),P(T_{m})\right) + \\
             &\phantom{=}~~\sum_{m} \sum_{n}  H\left(\hat{P}(J^n_{m} | T_{m} = t_{m}),P(J_{m}^{n} | T_{m} = t_{m})\right)
 \end{split}
\label{eq:loss2}
\end{equation}
\noindent
To model the ground truth distribution $\hat{P}$, we let $\hat{P}(T_{m})=0$ if $t_{m}=0$ and $\hat{P}(T_{m})=1$ if $t_{m}=1$, and specify the following likelihood models from the ground truth annotations,
\[
    \forall_{n} \forall_{m},~\hat{P}(J^n_m = j | T_{m} = t_m)=
\begin{cases}
    \mathcal{U}(j; 0,wh),& \text{if } t_m=0\\         \mathcal{G}(j; GT^{n}_{m}, \sigma^2\mathbb{I}),         & \text{if } t_m = 1
\end{cases}
 \label{eq:prob_map}
\]
\noindent
where $\mathcal{U}$ is a Uniform distribution in the interval 0 to $wh$ and $\mathcal{G}$ denotes a Gaussian distribution with mean $GT^{n}_{m}$ and covariance $ \sigma^2\mathbb{I}$ (\ie assuming a symmetric and diagonal covariance matrix). We use this Gaussian distribution to account for the inaccuracies in the ground truth annotations such that $\hat{P}(J^{n}_{m}|T_{m}=t_m)$ is a 2D probability map generated from the ground truth and which the network will try to estimate by optimizing Eq.~\eqref{eq:loss2}. In this work, we fix $\sigma^2 = 10$ for all experiments. That is, our network will optimize both the binary cross-entropy loss of each of the instruments as well as the sum of the pixel-wise probability map cross-entropy losses.

\subsection{Multi-Instrument Detector Network}
In order to provide a suitable network with the loss function of Eq.~\eqref{eq:loss2}, we modify and extend the U-Net~\cite{Ronneberger2015} architecture originally used for semantic segmentation.
Illustrated in Fig.~\ref{fig:unet}, the architecture uses down and up sampling stages, where each stage has a convolutional, a ReLU activation and a sampling layer. Here we use a total of 5 down and 5 up sampling stages and a single convolutional layer is used per stage to reduce the computational requirements. The number of features is doubled (down) or halved (up) per stage, starting with 64 features in the first convolutional layer. All convolutional kernels have a size of 3x3, except for the last layer where a 1x1 kernel is used. Batch normalization \cite{ioffe2015batch} is applied before every activation layer.
In order to provide output estimates $\forall (m,n), P(T_{m}), P(J_{m}^{n} | T_{m})$, we extend this architecture to do two things:
\begin{enumerate}
\item We create classification layers stemming from the lowest layer of the network by expanding it with a fully connected classification stage. The expansion is connected to the lowest layer in the network such that this layer learns to spatially encode the instruments. In particular this layer has one output per instrument which is activated with a sigmoid activation function to force a probabilistic output range. By doing so, we are effectively making the network provide estimates $P(T_{m})$.
\item Our network produces $M \times N$ maps of size $w \times h$ which correspond to each of the $P(J_{m}^{n} | T_{m}=1)$ likelihood distributions. Note that explicitly outputting $P(J_{m}^{n} | T_{m}=0)$ is unnecessary. Each output probability map ${P}(J_{m}^{n} | T_{m}=1)$ is normalized using a softmax function such that the joint position estimate of $GT^{n}_{m}$ is equal to the $\argmax_z P(J_{m}^{n} = z | T_{m} = 1)$.

\end{enumerate}
When combined with the loss function Eq.~\eqref{eq:loss2}, this network will train to both detect multiple instruments as well as estimate their joint parts. We implemented this network using the open source TensorFlow library \cite{Abadi2015} in Python\footnote{Code and models available at: \url{https://github.com/otl-artorg/instrument-pose}}.

\section{Experiments}\label{sec:experiments}
\textbf{Retinal Microsurgery.} We first evaluate our approach on the publicly available {\it in-vivo} retinal microsurgery instrument dataset~\cite{Sznitman2012}. The set contains 3 video sequences with 1171 images, each with a resolution of 640x480 pixels. Each image contains a single instrument with 4 annotated joints (start shaft, end shaft, left tip and right tip).
As in~\cite{Sznitman2012}, we trained our network on the first 50\% of all three sequences and evaluated the rest. Optimization of the network was performed with the Adam optimizer~\cite{kingma2014adam} using a batch size of 2 and an initial learning rate of $10^{-4}$. The network was trained for 10 epochs. Training and testing was performed on a Nvidia GTX 1080 GPU running at an inference rate of approximately 9 FPS.

\begin{figure}[t!]
\centering
\input{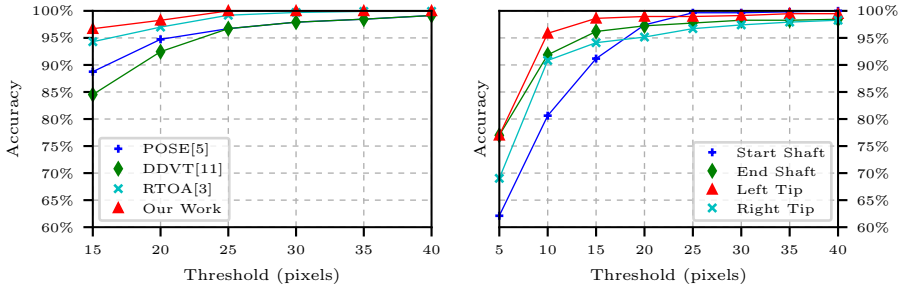}
\caption{Detection accuracy. (left) Percentage of correctly detected end of shaft joints as a function of the accuracy threshold. (right) Percentage of correctly detected joints.}
\label{fig:retina_results_1}
\end{figure}

\begin{figure}[t!]
\centering
\includegraphics[width=0.99\textwidth,height=170pt]{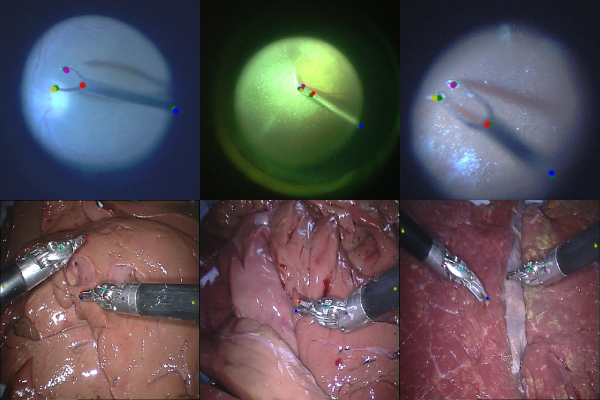}
\caption{Visual results on retinal microsurgery image sequences 1-3 ({\it top}) and laparoscopy sequences ({\it bottom}). The first two laparoscopy sequences contain claspers, whereas the right most contains a scissor and a clasper. The ground truths are denoted with green points. }
\label{fig:retina_results_2}
\end{figure}

The network was trained on three joints (left tip, right tip and end shaft) while only the end shaft joint was evaluated. Similar to~\cite{Sznitman2012,Rieke2015,Rieke2016}, we show the proportion of frames where the end shaft is correctly identified as a function of detection sensitivity. We show the performance of our approach as well as state-of-the-art detection and tracking methods in Fig.~\ref{fig:retina_results_1}. Our method achieves an accuracy of 96.7\% at a threshold radius of 15 pixels which outperforms the state-of-the-art of 94.3\%. The other two joints (left tip, right tip) achieve an accuracy of 98.3\% and 95.3\%, showing that the method is capable of learning all joint positions together with a high accuracy. The mean joint position errors are 5.1, 4.6 and 5.5 pixels. As the dataset includes 4 annotated joints, we  propose to also evaluate the performance for all joints and report in Fig.~\ref{fig:retina_results_1}(right) the accuracy of the joints after the network was trained with all joints using the same train-test data split. Overall, the performance is slightly lower than when training and evaluating with 3 joints because the 4th joint is the most difficult to detect due to blur and image noise. Fig.~\ref{fig:retina_results_2} depicts qualitative results of our approach and a video of all results can be found at \url{https://www.youtube.com/watch?v=ZigYQbGHQus}

\textbf{Robotic Laparoscopy.}
We also evaluated our approach on the MICCAI 2015 endoscopic vision challenge for laparoscopy instrument dataset tracking\footnote{\url{https://endovissub-instrument.grand-challenge.org/}}. The dataset includes 4 training and 6 testing video sequences. In total 3 different tools are visible in the sequences: left clasper, right clasper and left scissor which is only visible in the test set. The challenge data only includes a single annotated joint (extracted from the operating da Vinci\textregistered  robot) which is inaccurate in a large number of cases. For this reason, 5 joints (left tip, right tip, shaft point, end point, head point) per instrument in each image were manually labeled and then used instead \footnote{\url{https://github.com/surgical-vision/EndoVisPoseAnnotation}}. Images were resized to 640x512 pixels due to memory constraints when training the network. The training set consists of 940 images and the test set of 910 images. Presence of tools $T_{m}$ is given if a single joint is annotated.

We define the instruments $T_{1\ldots4}$ as left clasper, right clasper, left scissor and right scissor. To evaluate our approach, we propose two experiments: (1) Uses the same training and test data as in the original challenge, with an unknown tool in the test set. (2) We modified the training and test sets, such that the left scissor is also available during training by moving sequence 6 of the test set to the training set. By flipping the images in this sequence left-to-right, we augment our training data so to have the right scissor as well. Not only does this increase the complexity of the detection problem, but it also allows flipping data augmentation to be used.

\begin{figure}[t!]
\centering
\input{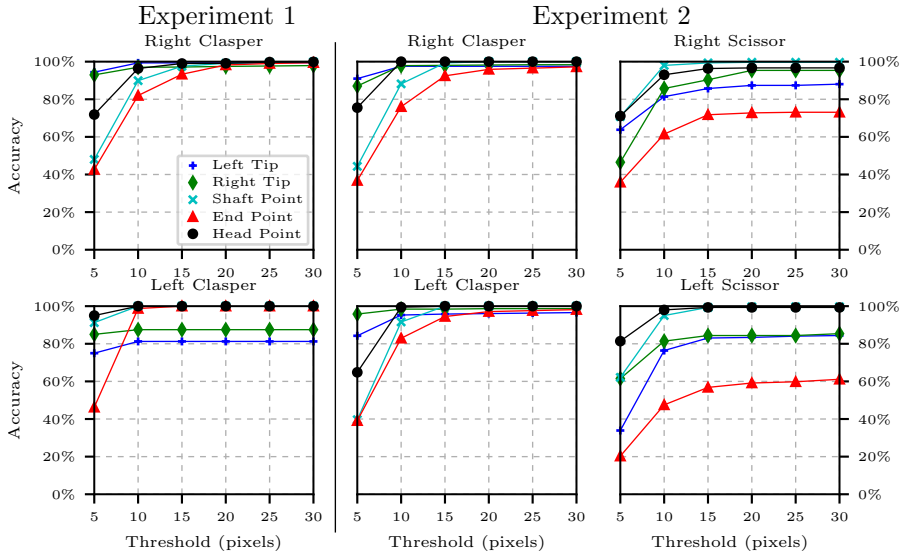}
\caption{Accuracy threshold curves: Left {\it Experiment 1} and right {\it Experiment 2}}
\label{fig:laparoscopy_results_1}
\end{figure}

{\it Experiment 1.} Using the original dataset, we first verified that the network can detect specific tools. As the left scissor has not been trained on, we expect this tool to be missed.
The training set was augmented using left-right and up-down flips. 
On the test set, only two images were wrongly classified, with an average detection rate of 99.9\% (right clasper 100\%, left clasper 99.89\%). Evaluation of the joint prediction accuracy was performed as with the microsurgery dataset and the results are illustrated in Fig.~\ref{fig:laparoscopy_results_1}(left). The accuracy is over 90\% at 15 pixels sensitivity on all joints except for the two tips on the left clasper.
The lower performance is explained by the left clasper only being visible in 40 frames, and to that the method fails on 7 images where the tool tips of both the left and right clasper are in the vicinity of each other or overlapping.

{\it Experiment 2.} Here the dataset was modified so that the right scissor is also visible in the training set by placing sequence 6 from the test set into the training set. The classification results of the instruments are: right clasper 100\%, left clasper 100\%, right scissor 99.78\% and left scissor 99.67\%. Fig.~\ref{fig:laparoscopy_results_1} (right) shows the results of all joint accuracies for this experiment. The accuracy of the left clasper tool is slightly improved compared to the previous experiment due to the increased augmented training size. However, the method still  fails on the same images as in {\it Experiment 1}. The scissors show similar results for both left and right, which is to be expected due to them being from the  same flipped images. Further, for the scissor results it is visible that one joint performs poorer than the rest. Upon visual inspection, we associate this performance drop to the inconsistency in our annotations and the joint not being visible in certain images. Given that our method assumes all joints are visible if a tool is present, detection failures occur when joints are occluded. Due to the increased input image size compared to the retinal microsurgery experiments, the inference rate is lower at around 6 FPS using the same hardware.

\section{Conclusion}\label{sec:conclusion}
We presented a deep learning based surgical instrument detector. The network collectively estimates joint positions and instrument presence using a combined loss function. Furthermore, the network obtains all predictions using a single feed-forward pass. We validated the method on two datasets, an {\it in-vivo} retinal microsurgery dataset and an {\it ex-vivo} laparoscopy set. Evaluations on the retinal microsurgery dataset showed state-of-the-art performance, outperforming even the current tracking methods. Our detector method is uninfluenced by previous estimations which is a key advantage over tracking solutions. The laparoscopy dataset showed that the method is capable of classifying instrument presence with a very high accuracy while jointly estimating the position of 20 joints. This points to our method being able to simultaneously count, estimate joint locations and classify whether instruments are visible in a single feed-forward pass.

{\small
\bibliographystyle{plain}
\bibliography{bibfiles}
}
\end{document}